\pdfoutput=1
\documentclass[11pt,a4paper]{article}

\PassOptionsToPackage{hyphens}{url}
\usepackage[hyperref]{emnlp2018}
\usepackage{times}
\usepackage{latexsym}
\usepackage{graphicx}
\usepackage{linguex}
\usepackage[plain]{fancyref}
\usepackage{textcomp}

\usepackage[utf8]{inputenc}
\usepackage[T1]{fontenc}
\usepackage[hyphens]{url}

\usepackage{booktabs}
\usepackage{multirow}
\usepackage{amssymb}

\usepackage{amsmath}
\usepackage{ifthen}
\usepackage{booktabs}   
\usepackage{microtype}  

\newcommand{\txt}[1]{\textrm{#1}}

\newcommand{\fixme}[2][]{%
    \ifthenelse{\equal{#1}{}}%
               {\textsl{[#2]}}%
               {\textsl{[#1: #2]}}%
    }%

\usepackage{numprint}
\npdecimalsign{.}
\npthousandsep{\ensuremath{\;}}
\npfourdigitnosep

\aclfinalcopy

\newcommand{\subs}{{\sc subs}{$_{\txt{full}}$}}
\newcommand{\subsH}[1]{{\sc subs}#1\emph{$_{H}$}}
\newcommand{\subsLM}[1]{{\sc subs}#1\emph{$_{LM}$}}

\DeclareMathOperator*{\softmax}{softmax}
\DeclareMathOperator*{\attention}{Att}
\DeclareMathOperator*{\multihead}{MH}
\DeclareMathOperator*{\feedforward}{FF}

\title{The MeMAD Submission to the WMT18 Multimodal Translation Task}

\author{%
Stig-Arne Grönroos \\
Aalto University \And
Benoit Huet \\
EURECOM \And
Mikko Kurimo \\
Aalto University \AND
Jorma Laaksonen \\
Aalto University \And
Bernard Merialdo \\
EURECOM \And
Phu Pham \\
Aalto University \AND
Mats Sjöberg \\
Aalto University \And
Umut Sulubacak \\
University of Helsinki \And
Jörg Tiedemann \\
University of Helsinki \AND
Raphael Troncy \\
EURECOM \And
Raúl Vázquez \\
University of Helsinki
}

\date{}

\begin{document}

\maketitle

\begin{abstract}

This paper describes the MeMAD project
entry to the WMT Multimodal Machine Translation Shared Task.

We propose adapting the Transformer neural machine translation (NMT)
architecture to a multi-modal setting.
In this paper, we also describe the preliminary experiments
with text-only translation systems leading us up to this choice.

We have the top scoring system for both English-to-German
and English-to-French,
according to the automatic metrics for \emph{flickr18}.

Our experiments show that
the effect of the visual features in our system is small.
Our largest gains come from the quality of the underlying text-only NMT system.
We find that appropriate use of additional data is effective.

\end{abstract}

\section{Introduction}
\label{sec:intro}

In multi-modal translation, the task is to translate from a source sentence
and the image that it describes, into a target sentence in another language.
As both automatic image captioning systems and crowd captioning efforts tend to mainly yield descriptions in English,
multi-modal translation can be useful for generating descriptions of images for languages other than English.
In the MeMAD project\footnote{\url{https://www.memad.eu/}}, multi-modal translation
is of interest for creating textual versions or descriptions of audio-visual content.
Conversion to text enables both indexing for multi-lingual image and video search,
and increased access to the audio-visual materials for visually impaired users.

We adapt%
\footnote{Our fork available from \url{https://github.com/Waino/OpenNMT-py/tree/develop_mmod}}
the Transformer \cite{vaswani2017attention} architecture
to use global image features extracted from Detectron,
a pre-trained object detection and localization neural network.
We use two additional training corpora: MS-COCO \cite{mscoco} and OpenSubtitles2018 \cite{opensubtitles}.
MS-COCO is multi-modal, but not multi-lingual.
We extended it to a synthetic multi-modal and multi-lingual training set.
OpenSubtitles is multi-lingual, but does not include associated images,
and was used as text-only training data.
This places our entry in the unconstrained category of the WMT shared task.
Details on the architecture used in this work can be found in Section \ref{sec:arch}.
Further details on the synthetic data are presented in Section \ref{sec:text_based}.
Data sets are summarized in Table \ref{tab:data}.

\begin{table}[t]
    \begin{center}
    {\small
    \begin{tabular}{lccccr}
    \toprule
    Data set       & images     & en         & de         & fr         & sentences \\ 
    \midrule
    Multi30k       & \checkmark & \checkmark & \checkmark & \checkmark & 29k     \\ 
    MS-COCO        & \checkmark & \checkmark & $+$        & $+$        & 616k    \\ 
    OpenSubtitles  &            & \checkmark & \checkmark & \checkmark & 23M/42M \\ 
                   & \multicolumn{5}{r}{1M, 3M, and 6M subsets used.} \\
    \bottomrule
    \end{tabular}}
    \end{center}
    \caption{Summary of data set sizes. 
        \checkmark means attribute is present in original data.
        $+$ means data set augmented in this work.}
    \label{tab:data}
\end{table}

\section{Experiment 1: Optimizing Text-Based Machine Translation}
\label{sec:text_based}

Our first aim was to select the text-based MT system
to base our multi-modal extensions on.
We tried a wide range of models,
but only include results with the two strongest systems:
Marian NMT with the \emph{amun} model \cite{mariannmt},
and OpenNMT \cite{opennmt} with the \emph{Transformer} model.

We also studied the effect of additional training data. Our initial experiments showed
that movie subtitles and their translations work rather well to augment the given training data.
Therefore, we included parallel subtitles from the OpenSubtitles2018 corpus to train better text-only MT models.
For these experiments, we apply the Marian amun model,
an attentional encoder-decoder model with bidirectional LSTM's on the encoder side.
In our first series of experiments, we observed that domain-tuning is very important when using Marian.
The domain-tuning was accomplished
by a second training step on in-domain data after training the model on the entire data set.
Table~\ref{tab:multidomain} shows the scores on development data.
We also tried decoding with an ensemble of three independent runs, which also pushed the performance a bit.

\begin{table}
\begin{center}
{\small
\begin{tabular}{lccc}
\toprule
\textsc{en-fr}         & flickr16 & flickr17 & mscoco17 \\
\midrule
multi30k               & 61.4     & 54.0     & 43.1     \\
\quad +\subs           & 53.7     & 48.9     & 47.0     \\
\quad\quad +domain-tuned      & 66.1     & 59.7     & \bf 51.7     \\
\quad\quad\quad +ensemble-of-3         & \bf 66.5     & \bf 60.2     & 51.6     \\
\bottomrule
\toprule
\textsc{en-de}         & flickr16 & flickr17 & mscoco17 \\
\midrule
multi30k               & 38.9     & 32.0     & 27.7     \\
\quad +\subs           & 41.3     & 34.1     & 31.3     \\
\quad\quad +domain-tuned      & 43.3     & 38.4     & 35.0     \\
\quad\quad\quad +ensemble-of-3         & \bf 43.9     & \bf 39.6     & \bf 37.0     \\
\bottomrule
\end{tabular}}
\end{center}
\caption{\label{tab:multidomain} Adding subtitle data and domain tuning for image caption translation (BLEU\% scores). All results with Marian Amun.} 
\end{table}

Furthermore, we tried to artificially increase the amount of in-domain data
by translating existing English image captions to German and French.
For this purpose, we used the large MS-COCO data set with its 100,000 images
that have five image captions each. We used our best multidomain model (see Table~\ref{tab:multidomain})
to translate all of those captions and used them as additional training data.
This procedure also transfers the knowledge learned by the multidomain model into the caption translations, which helps us to improve the coverage of the system with less out-of-domain data.
Hence, we filtered the large collection of translated movie subtitles
to a smaller portion of reliable sentence pairs (one million in the experiment we report)
and could train on a smaller data set with better results.

We experimented with two filtering methods.
Initially, we implemented a basic heuristic filter (\subsH{}),
and later we improved on this with a language model filter (\subsLM{}).
Both procedures consider each sentence pair, assign it a quality score,
and then select the highest scoring 1, 3, or 6 million pairs, discarding the rest.
The \subsH{} method counts terminal punctuation (\emph{`.', `...', `?', `!'}) in the source and target sentences,
initializing the score as the negative of the absolute value of the difference between these counts. Afterwards, it further
decrements the score by $1$ for each occurrence of terminal punctuation beyond the first in each of the sentences.
The \subsLM{} method first preprocesses the data by filtering samples by length and ratio of lengths,
applying a rule-based noise filter,
removing all characters not present in the Multi30k set,
and deduplicating samples.
Afterwards, target sentences in the
remaining pairs are scored using a character-based deep LSTM language model trained on the Multi30k data.
Both selection procedures are intended for noise filtering,
and \subsLM{} additionally acts as domain adaptation.
Table~\ref{tab:cocotrans} lists the scores we obtained on development data.

\setlength{\tabcolsep}{0.2em}
\begin{table}
  \begin{center}
    {\small
      \begin{tabular}{llccc}
        \toprule
          & \textsc{en-fr}       & flickr16 & flickr17  & mscoco17 \\
        \midrule
        A & \subsH{1M}$+$MS-COCO & 66.3     & 60.5      & 52.1     \\
        A & \quad +domain-tuned  & 66.8     & 60.6      & 52.0     \\
        A & \quad\quad +labels   & \bf 67.2 & 60.4      & 51.7     \\
        T & \subsLM{1M}+MS-COCO  & 66.9     & 60.3      & \bf 52.8 \\
        T & \quad +labels        & \bf 67.2 & \bf  60.9 & 52.7     \\
        \bottomrule
        \toprule
          & \textsc{en-de}       & flickr16 & flickr17  & mscoco17 \\
        \midrule
        A & \subsH{1M}$+$MS-COCO & 43.1     & 39.0      & 35.1     \\
        A & \quad +domain-tuned  & 43.9     & 39.4      & 35.8     \\
        A & \quad\quad +labels   & 43.2     & 39.3      & 34.3     \\
        T & \subsLM{1M}+MS-COCO  & \bf 44.4 & 39.4      & 35.0     \\
        T & \quad +labels        & 44.1     & \bf 39.8  & \bf 36.5 \\
        \bottomrule
      \end{tabular}}
  \end{center}
  \caption{\label{tab:cocotrans} Using automatically translated image captions and domain labels (BLEU\% scores).
    A is short for Amun, T for Transformer.}
\end{table}
\setlength{\tabcolsep}{0.5em}

To make a distinction between automatically translated captions,
subtitle translations and human-translated image captions, we also introduced domain labels
that we added as special tokens to the beginning of the input sequence. In this way,
the model can use explicit information about the domain when deciding how to translate given input.
However, the effect of such labels is not consistent between systems.
For Marian amun, the effect is negligible as we can see in Table~\ref{tab:cocotrans}.
For the Transformer, domain labels had little effect on BLEU 
but were clearly beneficial according to chrF-1.0.

\subsection{Preprocessing of textual data}

The final preprocessing pipeline for the textual data consisted of
lowercasing, tokenizing using Moses,
fixing double-encoded entities and other encoding problems,
and normalizing punctuation.
For the OpenSubtitles data we additionally used the \subsLM{} subset selection.

Subword decoding has become popular in NMT.
Careful choice of translation units
is especially important as one of the target languages of our system is German,
a morphologically rich language.
We trained a shared 50k subword vocabulary using Byte Pair Encoding (BPE) \cite{sennrich2015neural}.
To produce a balanced multi-lingual segmentation, the following procedure was used:
First, word counts were calculated individually for
English and each of the 3 target languages Czech\footnote{Czech was later dropped as a target language due to time constraints.}, French and German.
The counts were normalized to equalize the sum of the counts for each language.
This avoided imbalance in the amount of data skewing the segmentation in favor of some language.
Segmentation boundaries around hyphens were forced, overriding the BPE.

Multi-lingual translation with target-language tag 
was done following \newcite{johnson2016google}.
A special token, e.g. <TO\_DE> to mark German as the target language,
was prefixed to each paired English source sentence.

\section{Experiment 2: Adding Automatic Image Captions}

Our first attempt to add multi-modal information to the translation model includes the incorporation of automatically created image captions in a purely text-based translation engine. For this, we generated five English captions for each of the images in the provided training and test data.  This was done by using our in-house captioning system \cite{Shetty2018}.
The image captioning system uses a 2-layer LSTM with residual connections to generate captions based on scene context and object location descriptors, in addition to standard CNN-based features.
The model was trained with the MS-COCO training data and used to be state of the art in the COCO leaderboard\footnote{\url{https://competitions.codalab.org/competitions/3221}} in Spring 2016.
The beam search size was set to five.

We tried two models for the integration of those captions: (1) a dual attention multi-source model
that adds another input sequence with its own decoder attention and (2) a concatenation model
that adds auto captions at the end of the original input string separated by a special token.
In the second model, attention takes care of learning how to use the additional information
and previous work has shown that this, indeed, is possible \cite{niehues2016pre,ostling2017helsinki}. For both models,
we applied Marian NMT that already includes a working implementation of dual attention translations.
Table~\ref{tab:autocap} summarizes the scores on the three development test sets for English-French and English-German.

\setlength{\tabcolsep}{0.22em}
\begin{table}
  \small
  \begin{center}
    {\small
      \begin{tabular}{lccc}
        \toprule
        \textsc{en-fr}                  & flickr16 & flickr17 & mscoco17 \\
        \midrule
        multi30k                        & 61.4     & 54.0     & 43.1     \\
        \quad +autocap (dual attn.)     & 60.9     & 52.9     & 43.3     \\
        \quad +autocap 1   (concat)     & 61.7     & 53.7     & 43.9     \\
        \quad +autocap 1-5 (concat)     & \bf 62.2 & \bf 54.4 & \bf 44.1 \\
        \bottomrule
        \toprule
        \textsc{en-de}                  & flickr16 & flickr17 & mscoco17 \\
        \midrule
        multi30k                        & 38.9     & 32.0     & 27.7     \\
        \quad +autocap (dual attn.)     & 37.8     & 30.2     & 27.0     \\
        \quad +autocap 1   (concat)     & 39.7     & \bf 32.2 & \bf 28.8 \\
        \quad +autocap 1-5 (concat)     & \bf 39.9 & 32.0     & 28.7     \\
        \bottomrule
      \end{tabular}}
  \end{center}
  \caption{\label{tab:autocap} Adding automatic image captions (only the best one or all 5). The table shows BLEU scores in \%. All results with Marian Amun.}
\end{table}
\setlength{\tabcolsep}{0.5em}

\begin{figure}[h]
\flushright
\includegraphics[width= 1\hsize]{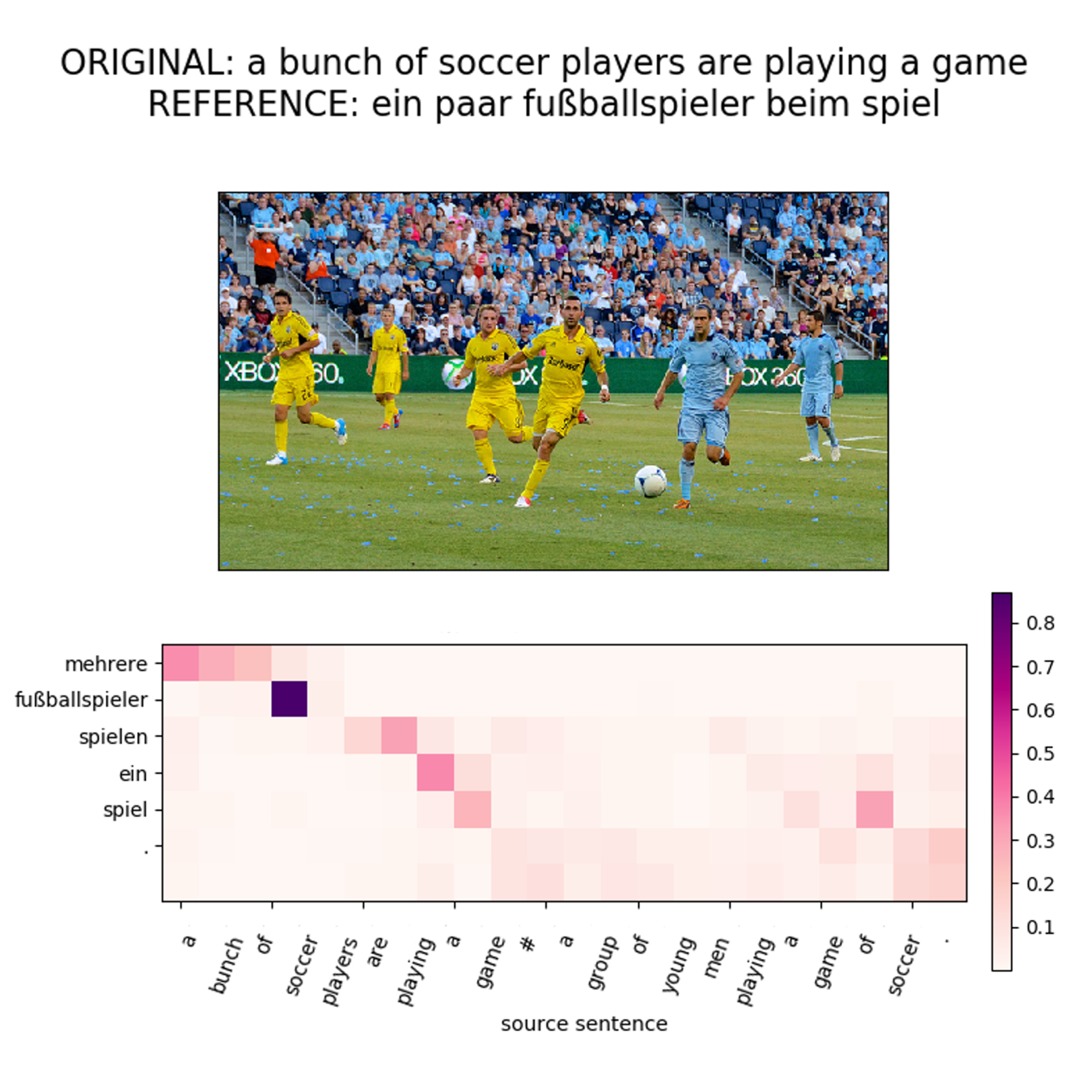} 
\caption{Attention layer visualization for an example where at least one of the attention weights for the last part of the sentence, which corresponds to the automatically generated captions, obtains a value above 0.3}
 \label{fig:autocaps}
\end{figure}

We can see that the dual attention model does not work at all and the scores slightly drop.
The concatenation approach works better probably because
the common attention model learns interactions between the different types of input.
However, the improvements are small if any and the model basically learns to ignore
the auto captions, which are often very different from the original input.
The attention pattern in the example of Figure~\ref{fig:autocaps} shows
one of the very rare cases where we observe at least some attention to the automatic captions.

\section{Experiment 3: Multi-modal Transformer}

One benefit of NMT,
in addition to its strong performance,
is its flexibility in enabling different information sources to be merged.
Different strategies to include image features both on the encoder and decoder side have been explored.
We are inspired by the recent success of the Transformer architecture
to adapt some of these strategies for use with the Transformer.

Recurrent neural networks start their processing from some \textbf{initial hidden state}.
Normally, a zero vector or a learned parameter vector is used,
but the initial hidden state is also a natural location to introduce additional context e.g. from other modalities.
Initializing can be applied in either the encoder (IMG$_{E}$) or decoder (IMG$_{D}$) \cite{calixto2017dcu}.
These approaches are not directly applicable to the Transformer,
as it is not a recurrent model, and lacks a comparable initial hidden state.

\textbf{Double attention} is another popular choice, used by e.g. \newcite{caglayan2017lium}. 
In this approach, two attention mechanisms are used, one for each modality.
The attentions can be separate or hierarchical.
While it would be possible to use double attention with the Transformer,
we did not explore it in this work.
The multiple multi-head attention mechanisms in the Transformer
leave open many challenges in how this integration would be done.

\textbf{Multi-task learning} has also been used,
e.g. in the Imagination model \cite{elliott2017imagination},
where the auxiliary task consists of reconstructing the visual features from the source encoding.
Imagination could also have been used with the Transformer,
but we did not explore it in this work.

The \textbf{source sequence} itself is also a possible location for including the visual information.
In the IMG$_{W}$ approach, 
the visual features are encoded as a pseudo-word embedding 
concatenated to the word embeddings of the source sentence.
When the encoder is a bidirectional recurrent network, as in \newcite{calixto2017dcu},
it is beneficial to add the pseudo-word 
both at the beginning and the end
to make it available for both encoder directions.
This is unnecessary in the Transformer,
as it has equal access to all parts of the source in the deeper layers of the encoder.
Therefore, we add the pseudo-word only to the beginning of the sequence.
We use an affine projection of the image features $V \in \mathbb{R}^{80}$
into a pseudo-word embedding $x_{I} \in \mathbb{R}^{512}$
\[x_{I} = W_{src} \cdot V + b_{I}.\]

In the LIUM \emph{trg-mul} \cite{caglayan2017lium},
the \textbf{target embeddings} and visual features are interacted through elementwise multiplication.
\[y'_{j} = y_{j} \odot \tanh(W_{mul}^{dec} \cdot V)\]
Our initial gating approach resembles \emph{trg-mul}.

\subsection{Architecture}
\label{sec:arch}

The baseline NMT for this experiment is the OpenNMT implementation of the Transformer.
It is an encoder-decoder NMT system
using the Transformer architecture \cite{vaswani2017attention} for both the encoder and decoder side.
The Transformer is a deep, non-recurrent network for processing variable-length sequences.
A Transformer is a stack of layers, consisting of two types of sub-layer:
multi-head (MH) attention (Att) sub-layers and feed-forward (FF) sub-layers:
\begin{align}
\attention(Q, K, V) & = \softmax(\frac{QK^T}{\sqrt{d_{k}}})V \nonumber\\
a_{i}        &      = \attention(QW_{i}^{Q}, KW_{i}^{K}, VW_{i}^{V}) \nonumber\\
\multihead(Q, K, V) & = [a_{1};\mathellipsis;a_{h}]W^{O} \nonumber\\
\feedforward(x)     & = \max(0, xW_{1} + b_{1})W_{2} + b_{2}
\end{align}
where $Q$ is the input query,
$K$ is the key,
and $V$ the attended values.
Each sub-layer is individually wrapped in a residual connection and layer normalization.

When used in translation, Transformer layers are stacked into an encoder-decoder structure.
In the encoder, the layer consists of a self-attention sub-layer followed by a FF sub-layer.
In self-attention, the output of the previous layer is used as queries, keys and values $Q = K = V$.
In the decoder, a third context attention sub-layer is inserted between the self-attention and the FF.
In context attention, $Q$ is again the output of the previous layer,
but $K = V$ is the output of the encoder stack.
The decoder self-attention is also masked to prevent access to future information.
Sinusoidal position encoding makes word order information available.

\textbf{Decoder gate}.
Our first approach is inspired by \emph{trg-mul}.
A gating layer is introduced to modify the pre-softmax prediction distribution.
This allows visual features to directly suppress a part of the output vocabulary.
The probability of correctly translating a source word with visually resolvable ambiguity
can be increased by suppressing the unwanted choices.

At each timestep the decoder output $s_{j}$ is projected to an unnormalized distribution over the target vocabulary.
\[ y_{j} = W \cdot s_{j} + b \]
Before normalizing the distribution using a softmax layer, a gating layer can be added.
\begin{align}
g  &= \sigma(W_{gate}^{dec} \cdot V + b_{gate}^{dec}) \nonumber\\
y'_{j} &= y_{j} \odot g
\end{align}

Preliminary experiments showed that gating based on only the visual features did not work.
Suppressing the same subword units during the entire decoding of the sentence was too disruptive.
We addressed this by using the decoder hidden state as additional input to control the gate.
This causes the vocabulary suppression to be time dependent.
\begin{align}
g_{j}  &= \sigma(U_{gate}^{dec} \cdot s_{j} + W_{gate}^{dec} \cdot V + b_{gate}^{dec}) \nonumber\\
\end{align}

\textbf{Encoder gate}.
The same gating procedure can also be applied to the output of the encoder.
When using the encoder gate, 
the encoded source sentence is disambiguated,
instead of suppressing part of the output vocabulary.
\begin{align}
g_{i}  &= \sigma(U_{gate}^{enc} \cdot h_{i} + W_{gate}^{enc} \cdot V + b_{gate}^{enc}) \nonumber\\
h'_{i} &= h_{i} \odot g_{i}
\end{align}

The gate biases $b_{gate}^{dec}$ and $b_{gate}^{enc}$ should be initialized to positive values,
to start training with the gates opened.
We also tried combining both forms of gating.

\begin{table}
\begin{center}
{\small
\begin{tabular}{lccc}
\toprule
\sc{en-fr}                          &     flickr16 &     flickr17 & mscoco17\\
\midrule
IMG$_{W}$            & \em 68.30    & \bf 62.45    &     52.86       \\
enc-gate             &     68.01    &     61.38    & \bf 53.40       \\ 
dec-gate             &     67.99    &     61.53    &     52.38       \\
enc-gate + dec-gate  & \bf 68.58    & \em 62.14    & \em 52.98       \\           
\bottomrule
\toprule
\sc{en-de}                          & flickr16     & flickr17     & mscoco17\\
\midrule
IMG$_{W}$            & \em 45.09    &     40.81    &     36.94    \\
enc-gate             &     44.75    & \bf 41.44    & \bf 37.76    \\
dec-gate             & \bf 45.21    &     40.79    &     36.47    \\
enc-gate + dec-gate  &     44.91    & \em 41.06    & \em 37.40    \\               
\bottomrule
\end{tabular}}
\end{center}
\caption{\label{tab:strategies}
Comparison of strategies for integrating visual information (BLEU\% scores).
All results using Transformer, Multi30k+MS-COCO+\subsLM{3M}, Detectron mask surface, and domain labeling.
}
\end{table}

\setlength{\tabcolsep}{0.23em}
\begin{table}
\begin{center}
{\small
\begin{tabular}{lccc}
\toprule
\sc{en-fr}                          &     flickr16 &     flickr17 & mscoco17\\
\midrule
\subsLM{3M} detectron               &     68.30    &     62.45    &     52.86 \\
\quad $+$ensemble-of-3                     &     68.72    &     62.70    &     53.06 \\
\quad\quad $-$visual features         & \bf 68.74    & \bf 62.71    &     53.14  \\
\quad $-$MS-COCO                      &     67.13    &     61.17    & \bf 53.34 \\
\quad $-$multi-lingual                &     68.21    &     61.99    &     52.40 \\
\subsLM{6M} detectron               &     68.29    &     61.73    &     53.05 \\
\subsLM{3M} gn2048                  &     67.74    &     61.78    &     52.76 \\
\subsLM{3M} text-only               &     67.72    &     61.75    &     53.02 \\
\bottomrule
\toprule
\sc{en-de}                          & flickr16     & flickr17     & mscoco17\\
\midrule
\subsLM{3M} detectron               &     45.09    &     40.81    &     36.94 \\
\quad $+$ensemble-of-3                     &     45.52    & \bf 41.84    & \bf 37.49 \\
\quad\quad $-$visual features         & \bf 45.59    &     41.75    &     37.43 \\
\quad $-$MS-COCO                      &     45.11    &     40.52    &     36.47 \\
\quad $-$multi-lingual                &     44.95    &     40.09    &     35.28 \\
\subsLM{6M} detectron               &     45.50    &     41.01    &     36.81 \\
\subsLM{3M} gn2048                  &     45.38    &     40.07    &     36.82 \\
\subsLM{3M} text-only               &     44.87    &     41.27    &     36.59 \\
\quad $+$multi-modal finetune       &     44.56    &     41.61    &     36.93 \\
\bottomrule
\end{tabular}}
\end{center}
\caption{\label{tab:multimodal} 
Ablation experiments (BLEU\% scores).
The row \subsLM{3M} \emph{detectron} shows our best single model.
Individual components or data choices are varied one by one.
$+$ stands for adding a component, and
$-$ for removing a component or data set.
Multiple modifications are indicated by increasing the indentation.}
\end{table}
\setlength{\tabcolsep}{0.5em}

\subsection{Visual feature selection}
\label{sec:visual_features}

Image feature selection was performed using the LIUM-CVC translation system~\cite{caglayan2017lium} training on the WMT18 training data, and evaluating on the \emph{flickr16, flickr17} and \emph{mscoco17} data sets.
This setup is different from our final NMT architecture as the visual feature selection stage was performed at an earlier phase of our experiments.
However, the LIUM-CVC setup without training set expansion was also faster to train which enabled a more extensive feature selection process.

We experimented with a set of state-of-the-art visual features, described below.

\textbf{CNN-based features} are 2048-dimensional feature vectors produced by applying reverse spatial pyramid pooling on features extracted from the 5$^{th}$ Inception module of the pre-trained GoogLeNet~\cite{szegedy2015going}.
For a more detailed description, see~\cite{Shetty2018}. 
These features are referred to as gn2048 in Table~\ref{tab:multimodal}.

\textbf{Scene-type features} 
are 397-dimensional feature vectors representing the association score of an image to each of the scene types in SUN397~\cite{xiao2010sun}. Each association score is determined by a separate Radial Basis Function Support Vector Machine (RBF-SVM) classifier trained from pre-trained GoogLeNet CNN features~\cite{Shetty2018}.

\textbf{Action-type features} are 40-dimensional feature vectors created with RBF-SVM classifiers similarly to the scene-type features, but using the Stanford 40 Actions dataset~\cite{conf/iccv/YaoJKLGF11} for training the classifiers.
Pre-trained GoogLeNet CNN features~\cite{szegedy2015going} were again used as the first-stage visual descriptors.

\textbf{Object-type and location features} are generated using the Detectron software\footnote{\url{https://github.com/facebookresearch/Detectron}} which implements Mask R-CNN~\cite{he2017mask} with ResNeXt-152~\cite{xie2017aggregated} features.
Mask R-CNN is an extension of Faster R-CNN object detection and localization~\cite{ren2015faster} that also generates a segmentation mask for each of the detected objects.
We generated an 80-dimensional \emph{mask surface} feature vector by expressing the image surface area covered by each of the MS-COCO classes based on the detected masks.

We found that the Detectron mask surface resulted in the best BLEU scores in all evaluation data sets for improving the German translations.
Only for \emph{mscoco17} the results could be slightly improved with a fusion of mask surface and the SUN 397 scene-type feature.
For French, the results were more varied, but we focused on improving the German translation results as those were poorer overall.
We experimented with different ways of introducing the image features into the translation model implemented in LIUM-CVC, and found as in~\cite{caglayan2017lium}, that \emph{trg-mul} worked best overall.

Later we learned that the \emph{mscoco17} test set has some overlap with the COCO 2017 training set, which was used to train the Detectron models.
Thus, the results on that test set may not be entirely reliable.
However, we still feel confident in our conclusions as they are also confirmed by the \emph{flickr16} and \emph{flickr17} test sets.

\subsection{Training}
\label{sec:train}

\begin{figure}[h]
\centering
\includegraphics[width=0.93\hsize]{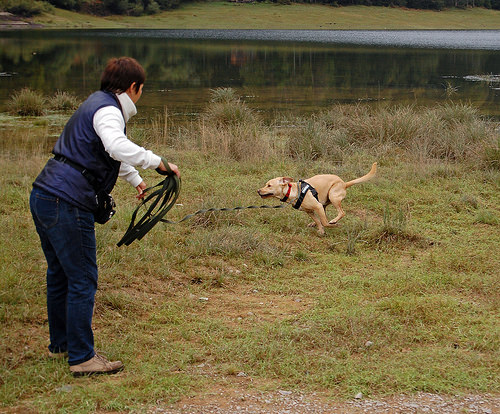} \caption{Image 117 was translated correctly as feminine ``eine besitzerin steht still und ihr brauner hund rennt auf sie zu .'' when not using the image features, but as masculine ``ein besitzer \textellipsis'' when using them. The English text contains the word ``her''. The person in the image has short hair and is wearing pants.}
 \label{fig:blinded}
\end{figure}

We use the following parameters for the network:%
\footnote{Parameters were chosen following the OpenNMT FAQ
\url{http://opennmt.net/OpenNMT-py/FAQ.html\#how-do-i-use-the-transformer-model}}
6 Transformer layers in both encoder and decoder,
512-dimensional word embeddings and hidden states,
dropout 0.1, batch size 4096 tokens, label smoothing 0.1,
Adam with initial learning rate 2 and $\beta_{2}$ 0.998.

For decoding, we use an ensemble procedure,
in which 
the predictions of 3 independently trained models
are combined by averaging after the softmax layer
to compute combined prediction.

We evaluate the systems using uncased BLEU using multibleu.
During tuning, we also used characterF \cite{popovic2015chrf} with $\beta$ set to 1.0.

There are no images paired with the sentences in OpenSubtitles.
When using OpenSubtitles in training multi-modal models,
we feed in the mean vector of all visual features in the training data
as a dummy visual feature.

\subsection{Results}
\label{sec:results}
Based on the previous experiments,
we chose the Transformer architecture,
Multi30k+MS-COCO+\subsLM{3M} data sets,
Detectron mask surface visual features,
and domain labeling.

Table~\ref{tab:strategies} shows the BLEU scores
for this configuration with different ways of integrating the visual features.
The results are inconclusive.
The ranking according to chrF-1.0 was not any clearer.
Considering the results as a whole and the simplicity of the method,
we chose IMG$_{W}$ going forward.

Table~\ref{tab:multimodal} shows results of ablation experiments
removing or modifying one component or data choice at a time,
and results when using ensemble decoding.
Using ensemble decoding gave a consistent but small improvement.
Multi-lingual models were clearly better than mono-lingual models.
For French, 6M sentences of subtitle data gave worse results than 3M.

We experimented with adding multi-modality to a pre-trained text-only system using a fine tuning approach.
In the fine tuning phase, a \emph{dec-gate} gating layer was added to the network.
The parameters of the main network were frozen, allowing only the added gating layer to be trained.
Despite the freezing, the network was still able to unlearn most of the benefits of the additional text-only data.
It appears that the output vocabulary was reduced back towards the vocabulary seen in the multi-modal training set.
When the experiment was repeated so that the finetuning phase included the text-only data,
the performance returned to approximately the same level as without tuning 
(+multi-modal finetune row in Table~\ref{tab:multimodal}).

To explore the effect of the visual features on the translation of our final model,
we performed an experiment where we retranslated using the ensemble while ``blinding'' the model.
Instead of feeding in the actual visual features for the sentence,
we used the mean vector of all visual features in the training data.
The results are marked \emph{-visual features} in Table~\ref{tab:multimodal}.
The resulting differences in the translated sentences were small,
and mostly consisted of minor variations in word order.
BLEU scores for French were surprisingly slightly improved by this procedure.
We did not find clear examples of successful disambiguation.
Figure~\ref{fig:blinded} shows one example of a detrimental use of visual features.

It is possible that adding to the training data
forward translations of MS-COCO captions from a text-only translation system introduced a biasing effect.
If there is translational ambiguity that should be resolved using the image,
the text-only system will not be able to resolve it correctly,
instead likely yielding the word that is most frequent in that textual context.
Using such data for training a multi-modal system might bias it towards ignoring the image.

On this year's \emph{flickr18} test set,
our system scores 38.54 BLEU for English-to-German
and 44.11 BLEU for English-to-French.

\section{Conclusions}
\label{sec:concl}

Although we saw an improvement from incorporating multi-modal information,
the improvement is modest.
The largest differences in quality between the systems we experimented with
can be attributed to the quality of the underlying text-only NMT system.

We found the amount of in-domain training data 
and multi-modal training data to be of great importance. 
The synthetic MS-COCO data was still beneficial,
despite being forward translated,
and the visual features being over-confident due to being extracted from a part of the image classifier training data.

Even after expansion with synthetic data,
the available multi-modal data is dwarfed by the amount of text-only data.
We found that movie subtitles worked well for this purpose.
When adding text-only data, domain adaptation was important, 
and increasing the size of the selection met with diminishing returns.

Current methods do not fully address the problem of how to efficiently learn
from both large text-only data and small multi-modal data simultaneously.
We experimented with a fine tuning approach to this problem, without success.

Although the effect of the multi-modal information was modest,
our system still had the highest performance of the task participants
for the English-to-German and English-to-French language pairs,
with absolute differences of +6.0 and +3.5 BLEU\%, respectively.

\section*{Acknowledgments}
\label{sec:ack}
This work has been supported by the European Union's Horizon 2020 Research and Innovation Programme under Grant Agreement No 780069, and by the Academy of Finland in the project 313988.
In addition the Finnish IT Center for Science (CSC) provided computational resources.
We would also like to acknowledge the support by NVIDIA and their GPU grant.

\bibliography{main}
\bibliographystyle{emnlp_natbib}

\end{document}